\pdfoutput=1
\documentclass[11pt]{article}

\usepackage{acl}

\usepackage{times}
\usepackage{latexsym}
\usepackage{graphicx}
\usepackage[textsize=tiny]{todonotes}
\usepackage{booktabs}       
\usepackage{listings}
\usepackage{acronym}
\usepackage{tabularx}
\usepackage{multirow}

\usepackage[T1]{fontenc}

\usepackage{pythontex} 
\usepackage{xcolor}
\definecolor{codegreen}{rgb}{0,0.6,0}

\usepackage[utf8]{inputenc}

\usepackage{microtype}

\acrodef{S2U}{Speech-to-Units}
\acrodef{U2S}{Units-to-Speech}
\acrodef{U2U}{Units-to-Units}
\acrodef{uLM}{Unit-Language-Model}
\acrodef{ASR}{Automatic Speech Recognition}
\acrodef{TTS}{Text-to-Speech}
\acrodef{WER}{Word Error Rate}
\acrodef{CER}{Character Error Rate}
\acrodef{CPC}{Contrastive Predictive Coding}
\acrodef{NLP}{Natural Language Processing}

\usepackage{bm}
\usepackage{amssymb,amsmath,graphicx,bbm,color,booktabs}
\usepackage{amsopn}

\newcommand{\vx}{\bm{x}}               \newcommand{\xh}{\hat{x}}
               
\newcommand{\vz}{\bm{z}}       \newcommand{\vzh}{\hat{\bm{z}}}        

    \newcommand{\Xc}{\mathcal{X}}

\newcommand{\R}{\mathbb{R}}

\renewcommand{\eqref}[1]{Eq.~(\ref{#1})}

\def \z{{\mathbf z}}

\title{\texttt{textless-lib}: a Library for \\ 
 Textless Spoken Language Processing}

\makeatletter
\def\thanks#1{\protected@xdef\@thanks{\@thanks
        \protect\footnotetext{#1}}}
\makeatother

\author{
\normalsize
Eugene Kharitonov$^\bigstar$, Jade Copet$^\bigstar$, Kushal Lakhotia$^\blacktriangle$, Tu Anh Nguyen$^\bigstar$, Paden Tomasello$^\bigstar$\\ 
\normalsize
\textbf{Ann Lee$^\bigstar$, Ali Elkahky$^\bigstar$, Wei-Ning Hsu$^\bigstar$, Abdelrahman Mohamed$^\bigstar$, Emmanuel Dupoux$^{\bigstar\dagger}$, Yossi Adi$^\bigstar$}  \\
\normalsize
  $^\bigstar$ Meta AI Research, $^\dagger$ EHESS\\
\normalsize
  $^\blacktriangle$ Outreach\\
\small
  \texttt{\{kharitonov, jadecopet, adiyoss\}@fb.com}\\
}

\begin{document}
\maketitle

\begin{abstract}
Textless spoken language processing research aims to extend the applicability of standard NLP toolset onto spoken language and languages with few or no textual resources. In this paper, we introduce \texttt{textless-lib}, a PyTorch-based library aimed to facilitate research in this research area. We describe the building blocks that the library provides and demonstrate its usability by discuss three different use-case examples: (i) speaker probing, (ii) speech resynthesis and compression, and (iii) speech continuation. We believe that \texttt{textless-lib} substantially simplifies research the textless setting and will be handful not only for speech researchers but also for the NLP community at large. The code, documentation, and pre-trained models are available at \url{https://github.com/facebookresearch/textlesslib/}.


\end{abstract}

\vspace{-0.1cm}
\section{Introduction}
\label{sec:intro}
\vspace{-0.2cm}


Textless spoken language modeling \cite{Lakhotia2021} consists in jointly learning the acoustic and linguistic characteristics of a natural language from raw audio samples without access to textual supervision (e.g. lexicon or transcriptions).
This area of research has been made possible by converging progress in self-supervised speech representation learning~\cite{schneider2019wav2vec, Baevski2020, oord2018representation, Hsu2021, chorowski2021aligned, chen2021wavlm, chung2021w2v, wang2021unispeech, ao2021speecht5}, language modeling~\cite{peters-etal-2018-deep, devlin-etal-2019-bert, liu2019roberta, brown2020language, lewis-etal-2020-bart}, and speech synthesis~\cite{ren2019fastspeech, kumar2019melgan, yamamoto2020parallel, ren2020fastspeech, kong2020hifi, morrison2021chunked}.

\citet{Lakhotia2021} presented a Generative Spoken Language Modeling (GSLM) pipeline trained from raw audio, consisting in a speech encoder (converting speech to discrete units), a language model (based on units) and a decoder (converting units back to speech). These components enabled the generation of new speech by sampling units from the language model. 
\citet{Polyak2021} proposed an improved encoder/decoder working from disentangled quantized content and F0 units  
and showed how such a system could be used for efficient audio compression. \citet{Kharitonov2021} proposed a modified language model 
system capable of jointly modelling content units and F0 yielding expressive generations. 
Lastly, \citet{Kreuk2021} demonstrated that the language model can be replaced by a sequence to sequence model achieving the first high quality speech emotion conversion system (including laughter and yawning). 

\begin{figure}
    \centering
    \includegraphics[width=0.38\textwidth]{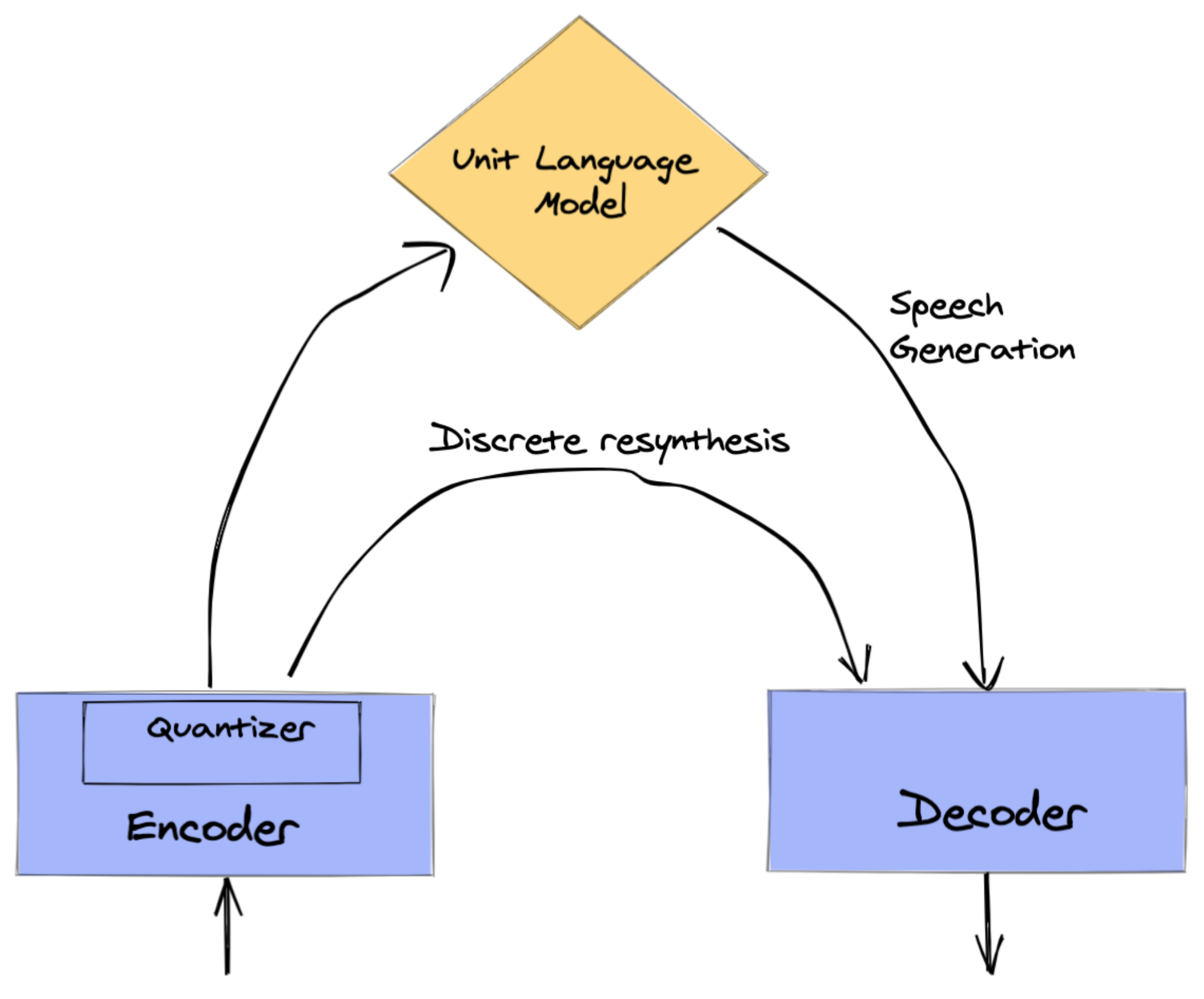}
    \caption{A visual description for textless modeling of spoken language. One can perform language modeling for speech continuations~\cite{Lakhotia2021} or a direct speech resynthesis~\cite{Polyak2021}.}
    \vspace{-0.2cm}
    \label{fig:pipeline}
\end{figure}

The textless approach has several advantages. First, it would be beneficial for the majority of the world's languages that do not have large textual resources or even a widely used standardized orthography (Swiss German, dialectal Arabic, Igbo, etc.). Despite being used by millions of people, these languages have little chance of being served by current text-based technology. Moreover, ``high-resource'' languages can benefit from such modeling where the oral and written forms are mismatched in terms of lexicon and syntax. Second, directly modeling spoken language from raw audio allows us to go beyond lexical content and also model linguistically relevant signals such as prosodic features, intonation, non-verbal vocalizations (e.g., laughter, yawning, etc.), speaker identity, etc. All of these are virtually absent in text. 

Although great progress has been made in modeling spoken language, it still requires domain expertise and involves a complicated setting. For instance, the official implementation of the GSLM pipeline~\citep{Lakhotia2021} consists of roughly four different launching scripts with a few dozens of checkpoints. Similarly, running the official implementation of~\citet{Polyak2021}, requires using four scripts from two different repositories.

In this work, we present \texttt{textless-lib}, a PyTorch library for textless spoken language processing. \texttt{textless-lib} makes the processing, encoding, modeling, and generating of speech as simple as possible. With a few lines of code, one can perform speech continuation, audio-book compression, representation analysis by probing, speech-to-speech translation, etc. We provide all the necessary building blocks, example pipelines, and example tasks. We believe such a simple to use API will encourage both speech and NLP communities to deepen and extend the research work on modeling spoken language without text and unlock potential future research directions.

\vspace{-0.2cm}
\section{Background}
\label{sec:background}
\vspace{-0.2cm}

Below we provide an overview of the common textless spoken language modeling pipeline. In a nutshell, such pipeline is usually comprised of: i) \ac{S2U} encoders that automatically discover discrete representations or units which can be used to encode speech into "pseudo-text"; ii) \ac{U2U} models that are used for units modeling. This can take a form as \ac{uLM} for speech continuation~\cite{Lakhotia2021, Kharitonov2021}, sequence-to-sequence models for speech emotion conversion~\cite{Kreuk2021} or translation tasks~\cite{Lee2021direct, Lee2021textless}; iii) \ac{U2S} models to reconstruct back the speech signals.

Alternatively, one could drop the \ac{U2U} component and perform a direct speech resynthesis~\cite{Polyak2021}. This can be used for speech compression, voice conversion, or developing a better understanding of the learned representation using probing methods. See Figure~\ref{fig:pipeline} for a visual description of the full system. We provide a detailed description for each of the above-mentioned components in the following subsections. 

\begin{table}[t!]
\centering
\resizebox{0.8\columnwidth}{!}{
\begin{tabular}{llc}
\toprule
\textbf{Type} & \textbf{Model} & \textbf{Dataset} \\
\midrule
\multirow{2}{*}{Encoders}       & HuBERT  & LS-960    \\
                              & CPC    &  LL-6k  \\ \midrule
\multirow{4}{*}{Quantizers}  & \multirow{4}{*}{k-means} & LS-960 w. 50 units\\
& & LS-960 w. 100 units \\ 
& & LS-960 w. 200 units \\ 
& & LS-960 w. 500 units \\
\midrule
F0 extract.         & YAAPT  & -    \\
\midrule
\multirow{2}{*}{Decoders}     & Tacotron2 & LJ Speech \\ 
                            & WaveGlow & LJ Speech \\
\bottomrule      
\end{tabular}
}
\caption{Summary of the pre-trained models provided in \texttt{textless-lib}. We denote LibriSpeech and LibriLight as LS-960 and LL-6k accordingly. All quantizers were trained on ``dev-clean'' partition of LibriSpeech.}
\vspace{-0.3cm}
\label{tab:models}
\end{table}

\vspace{-0.1cm}
\subsection{Speech to Units}
Consider the domain of audio samples as $\Xc \subset \R$. The representation for an audio waveform is therefore a sequence of samples $\vx = (x^1, \dots, x^T)$, where each $x^i \in \Xc$ for all $1 \leq t \leq T$. We denote the \ac{S2U} encoder as, 
$E(\vx) = \z$, 
where $\z = (\vz^{1}, \dots, \vz^{L})$ is a spectral representation of $\vx$ sampled at a lower frequency, each $\vz^{i}$ for $1 \le i \le L$ is a d-dimensional vector, and $L < T$. 

Next, as the representations obtained by $E$ are continuous, an additional quantization step is needed. We define a quantization function $Q$, which gets as input dense representations and outputs a sequence of discrete tokens corresponding to the inputs' quantized version. Formally, $Q(\z) = \vz_q$, where $\vz_q = (z_q^1, \dots, z_q^L)$ such that $z_q^i \in \{1, \dots, K\}$ and $K$ is the size of the vocabulary. After quantization one can either operate on the original discrete sequences (duped) or collapse repeated units (e.g., $0,0,1,1,2 \rightarrow 0,1,2$), we refer to such sequences as ``deduped''. Working with the deduped sequences simplifies modeling long sequences, however, the tempo information is lost.

\begin{figure*}[t!]
    \centering
    \includegraphics[width=0.83\textwidth]{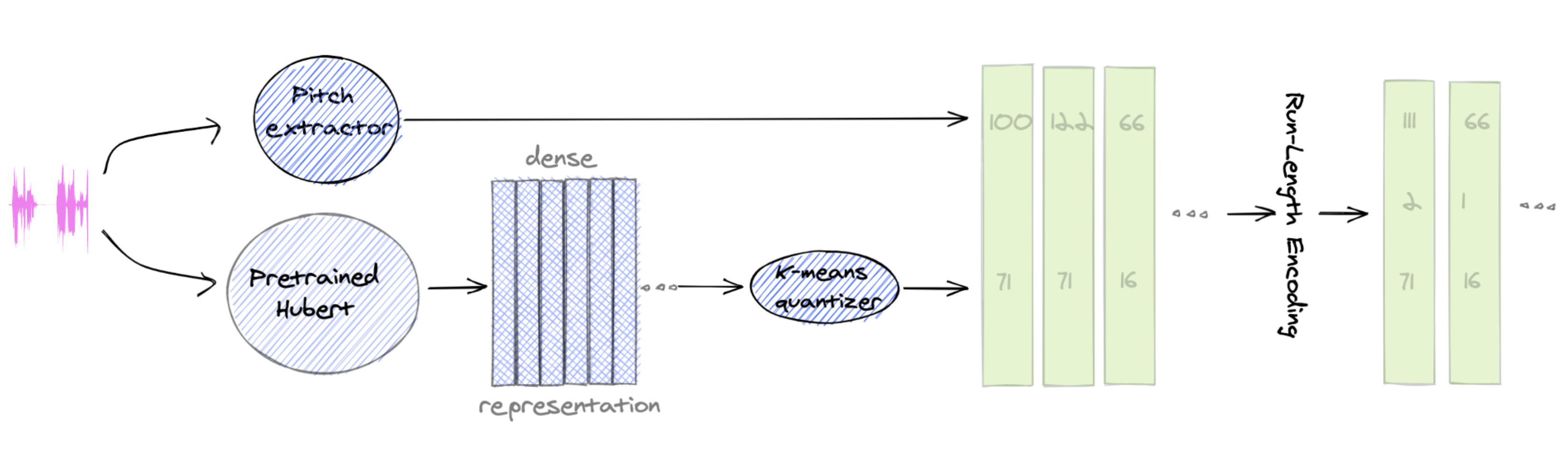}
    \caption{We represent speech as three aligned, synchronised streams: discrete pseudo-units, duration, and pitch.}
    \vspace{-0.2cm}
    \label{fig:preprocessing}
\end{figure*}

\vspace{-0.1cm}
\subsection{Units to Speech}
Converting a sequence of units to audio is akin to the \ac{TTS} problem, where we consider the discrete units as ``pseudo-text''. This can be solved by adopting a standard \ac{TTS} architecture. For instance, \citet{Lakhotia2021} trained a Tacotron2 model~\cite{Shen2018} to perform units to mel-spectrogram conversion followed by a WaveGlow~\cite{Prenger2019} neural vocoder for time-domain reconstruction. 

Formally, to reconstruct a time-domain speech signal from a sequence of discrete units, $\vz_q$ we define the composition as,
$V(G(\vz_q)) = \xh, $
where $G$ is a mel-spectrogram estimation module (e.g., Tacotron2), and $V$ is a phase vocoder module responsible for time-domain synthesis (e.g., WaveGlow). The input sequence $\vz_q$ can be either the original sequence or its deduped version. 

Interestingly, one can simplify the synthesis process when working with the duped unit sequences. As we have a direct mapping between the duped discrete unit sequence to the time domain signal (e.g., each unit corresponds to a 20ms window) one can remove $G$, and directly feed $\vz_q$ to $V$. This was successfully done in~\cite{Polyak2021} for speech resynthesis using the HiFi-GAN neural vocoder~\cite{kong2020hifi}. Alternatively, as suggested by~\cite{Kreuk2021, Lee2021textless} one can train a unit duration prediction model and use the predicted durations to inflate the sequence and feed the discrete sequence directly to $V$. 

\vspace{-0.2cm}
\subsection{Unit to Units}
\vspace{-0.2cm}
Equipped with the models to encode spoken language into discrete unit sequences and convert them back to speech samples, one can conveniently use the common \ac{NLP} architectures to model spoken language. 

Consider $M$ to be a sequence modeling function that gets as input a discrete unit sequence $\vz_q$ and outputs another discrete units sequence, denoted as $\vzh_q$. Generally, $\vzh_q$ can represent different generations, depending on the modeling task. For instance, \citet{Lakhotia2021, Kharitonov2021} set $M$ to be a Transformer language model and trained a generative spoken language model. Similarly, \citet{Kreuk2021} set $M$ to be a sequence-to-sequence model, hence can cast the emotion conversion problem as a translation task.\footnote{Examples are provided at \url{speechbot.github.io/}.}

\vspace{-0.2cm}
\section{Library Overview}
\label{sec:lib}
\vspace{-0.1cm}

In this section, we present the \texttt{textless-lib} library, intending to simplify future research on textless spoken language modeling. Additionally, the proposed package will remove the main barrier of processing and synthesizing speech, which requires domain expertise, for other language researchers (e.g., NLP researchers) who are interested in modeling spoken language, analyzing the learned representations, etc. 

To support the above, it is essential to provide the main building blocks described in Section~\ref{sec:background}, together with pre-trained models, with minimal coupling between them (a list of the supported pre-trained models can be seen on Table~\ref{tab:models}). This will allow researchers to flexibly use the provided pre-trained building blocks as well as develop new building blocks and use them anywhere in their pipeline. We decided to exclude both \ac{U2U} models as well as evaluation metrics from the core functionality of the library as we believe these models should be an example usage. There are plenty of ways to evaluate the overall pipeline~\cite{Lakhotia2021,dunbar2019zero,dunbar2020zero,Nguyen2020} as well as different ways to model the ``pseudo-text'' units~\cite{shi2021discretization, Kharitonov2021, Polyak2021, Kreuk2021, Lee2021direct}, hence including them as an integral part of \texttt{textless-lib} will make the library over complicated and hard to use. 

\vspace{-0.1cm}
\subsection{Interfaces}
The pipeline presented in Figure~\ref{fig:pipeline} hints a straightforward way to decouple elements of the library into two principal blocks: (i) encoding speech; and (ii) decoding speech, with the only inter-dependence being the format of the data in-between (e.g., vocabulary size). Such interfaces enable interesting mix-and-match combinations as well as conducting research on each component independently. We firstly present those two interfaces, then we discuss helpers for dataloading.

\begin{figure}[t!]
    \centering
    \includegraphics[width=0.5\textwidth]{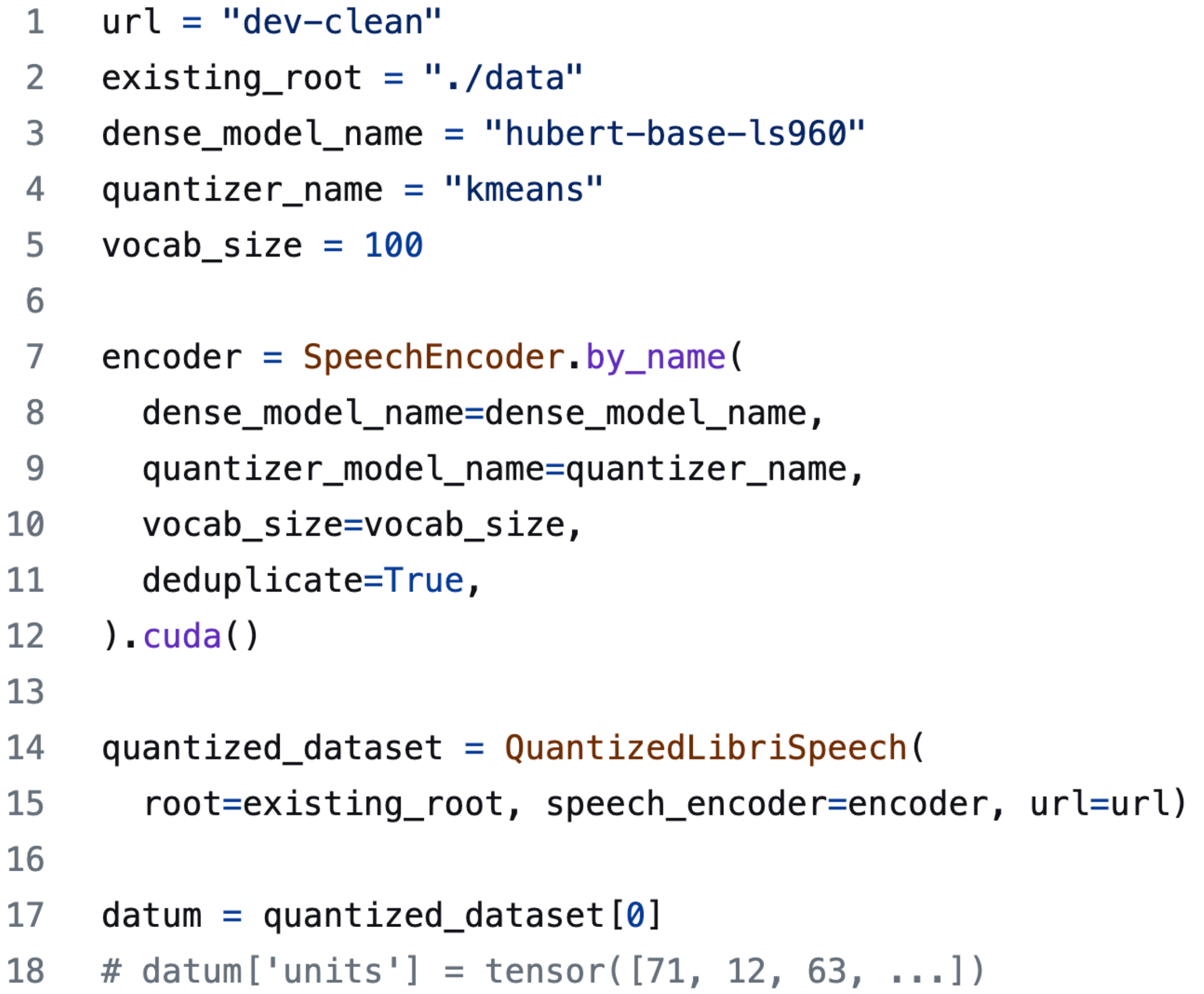}
    \caption{\texttt{textless-lib} provides an ``encoded'' view for standard datasets, such as LibriSpeech.}
    \label{fig:librispeech}
\end{figure}

{\noindent \bf{Encoders and Vocoders.}}
We denote the encoders as \texttt{SpeechEncoder}. These modules encompass all steps required to represent raw audio as discrete unit sequences (i.e., pseudo-text units and, optionally duration and pitch streams). 

\texttt{SpeechEncoder} obtains a dense vector representation from a given self-supervised model, discretizes the dense representation into units, extracts pitch, aligns it with the unit streams, and potentially, applies run-length encoding with per-frame pitch averaging. See Fig.~\ref{fig:preprocessing} for a visual description. 

For each sub-model, a user might choose to use a pre-trained model or provide a custom \texttt{torch.nn.Module} module instead. An example of the former is demonstrated in lines 7-12 in Figure~\ref{fig:librispeech}, in which a HuBERT model and a corresponding k-means codebook with a pre-defined $K$ (i.e., vocabulary size) are automatically retrieved. 

\begin{figure}[t!]
    \centering
    \includegraphics[width=0.5\textwidth]{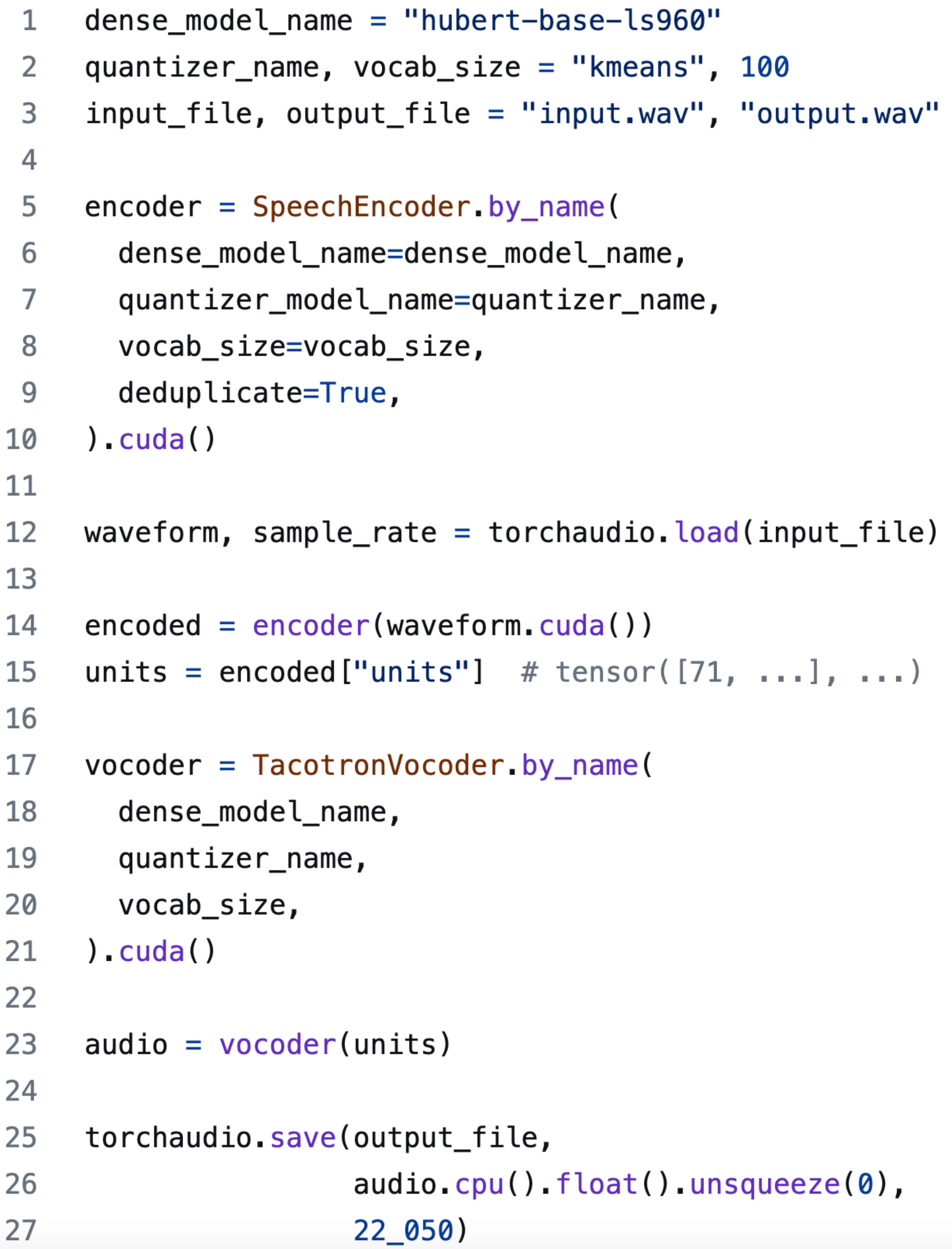}
    \caption{Fully functioning code for discrete audio resynthesis. An audio file is loaded, coverted into a sequence of pseudo-units and transformed back into audio with Tacotron2. The model setup code will download required checkpoints and cache them locally.\vspace{-0.3cm}}
    \label{fig:resynth_listing}
\end{figure}

Conversely, vocoders take as input a discretized sequence and convert it back to the audio domain. As with \texttt{SpeechEncoder}, we can retrieve a pre-trained model by setting the expected input specification (model, quantizer, and the size of the codebook), see Figure~\ref{fig:resynth_listing} lines 17-21. 

{\noindent \bf{Datasets, Dataloaders, and Preprocessing.}}
Apart from encoders and vocoders, in the \texttt{textless-lib} we provide several components aimed to simplify frequent data loading use-cases. First, we provide a set of standard datasets (e.g., LibriSpeech) wrapped to produce quantized representations (see~Fig.~\ref{fig:librispeech} lines 14-15). Those datasets are implemented via a \texttt{QuantizeDataset} wrapper which can be used to wrap any map-style PyTorch dataset, containing raw waveform data.

The \texttt{QuantizeDataset} runs an instance of a dense representation model, which can be computationally heavy (e.g., the HuBERT-base model has 7 convolutional layers and 12 Transformer layers). Unfortunately, such heavy preprocessing can starve the training loop. Hence, we provide two possible solutions: (i) as part of the \texttt{textless-lib} we provide a way to spread \texttt{QuantizeDataset} and \texttt{DataLoader} preprocessing workers (each with its copy of a dense model) across multiple GPUs, hence potentially balancing training and preprocessing across different devices; (ii) in cases where on-the-fly preprocessing is not required (e.g., there is no randomized data augmentation~\cite{Kharitonov2021b}), an alternative is to preprocess the entire dataset in advance. \texttt{textless-lib} provides a tool for preprocessing arbitrary sets of audio files into a stream of pseudo-unit tokens and, optionally, streams of per-frame tempo and F0 values, aligned to the token stream. The tool uses multi-GPU and multi-node parallelism to speed up the process.

\begin{table}[t!]
\centering
\resizebox{0.9\columnwidth}{!}{
\begin{tabular}{lccc}
\toprule
\textbf{Model} & \textbf{Quantized?} & \textbf{Vocab.~size }& \textbf{Accuracy} \\
\midrule
HuBERT & - & - & 0.99 \\
HuBERT & \checkmark & 50 & 0.11 \\
HuBERT & \checkmark & 100 & 0.19 \\
HuBERT & \checkmark & 200 &  0.29 \\
HuBERT & \checkmark & 500 &  0.48 \\
\midrule
CPC & - & - & 0.99 \\
CPC & \checkmark & 50 &  0.19 \\
CPC & \checkmark & 100 &  0.32  \\
CPC & \checkmark & 200 &  0.34 \\
CPC & \checkmark & 500 &  0.40 \\
\bottomrule      
\end{tabular}
}
    \caption{Speaker probing. Test accuracy 
    on predicting speaker based on HuBERT \& CPC representations. 
    }
    \vspace{-0.2cm}
    \label{tab:speaker_probing}
\end{table}

\vspace{-0.1cm}
\subsection{Pre-trained Models}
As part of \texttt{textless-lib} we provide several pre-trained models that proved to work best in prior work~\cite{Lakhotia2021,Polyak2021}. In future, we will maintain the list of the models to be aligned with state-of-the-art. 

{\noindent \bf{Dense representations.}} We support two dense representation models: (i) HuBERT base-960h model~\cite{Hsu2021} trained on LibriSpeech 960h dataset, with a framerate of 50 Hz; (ii) \ac{CPC} model~\cite{Riviere2020,oord2018representation} trained on the 6K hours subset from LibriLight~\cite{Kahn2020} with a framerate of 100 Hz. Both models provided the best overall performance according to~\cite{Lakhotia2021, Polyak2021}.

{\noindent \bf{Pitch extraction.}} Following~\citet{Polyak2021} we support F0 extraction using the YAAPT pitch extraction algorithm~\cite{Kasi2002}. 
We plan to include other F0 extraction models, e.g.~CREPE~\cite{kim2018crepe}. 

{\noindent \bf{Quantizers.}} With the \texttt{textless-lib} we provide several pre-trained quantization functions for both HuBERT and CPC dense models using a vocabulary sizes $K \in \{50, 100, 200, 500\}$. For the quantization function, we trained a k-means algorithm using the ``dev-clean'' part in the LibriSpeech dataset~\cite{Panayotov2015}.


{\noindent \bf{Pitch normalization.}} Following \citet{Kharitonov2021}, we applied per-speaker pitch normalization to reduce inter-speaker variability. For single speaker datasets, we do not perform F0 normalization and the span of pitch values is preserved. Under the \texttt{textless-lib} we provide two pitch-normalization methods: per-speaker and prefix-based. In the per-speaker normalization, we assume the mean F0 value per speaker is known in advance. While in the prefix-based normalization method a part of the audio is used to calculate the mean pitch. Those two options provide useful trade-offs. In the first case, we need to have a closed set of speakers but have a better precision while in the second we sacrifice quality but gain flexibility.

{\noindent \bf{Vocoder.}} In the initial release of the library, we provide Tacotron2 as a mel-spectrogram estimation module (i.e., the $G$ function) followed by WaveGlow~\cite{Prenger2019} neural vocoder (i.e., the $V$ function) as used by~\citet{Lakhotia2021}.\footnote{WaveGlow is used as a part of \texttt{TacotronVocoder}. Both Tacotron2 and WaveGlow were trained on LJ speech~\cite{ljspeech17}.} These operate on deduplicated pseudo-unit streams with vocabulary sizes of 50, 100, and 200. In a follow-up release, we aim to include HiFi-GAN-based vocoders similarly to~\citet{Polyak2021,Kharitonov2021}. We found those to generate better audio quality with higher computational performance. However, as described in Section~\ref{sec:background}, the main drawback of dropping $G$ and directly feeding the discrete units to $V$ is the need for a unit duration prediction model. We plan to include such models as well in the next release. 
\begin{table}[t!]
\centering
\resizebox{0.83\columnwidth}{!}{
\begin{tabular}{cccc}
\toprule
\textbf{Model} & \textbf{Vocab.~size }& \textbf{Bitrate}, bit/s & \textbf{WER} \\
\midrule
Topline & - & 512 $\cdot10^{3}$ & 2.2 \\
\midrule
HuBERT & 50  &  125.5 & 24.2 \\
HuBERT & 100 &  167.4 & 13.5\\
HuBERT & 200 & 210.6  & 7.9 \\
\bottomrule      
\end{tabular}}
\caption{Bitrate/ASR WER trade-off. 
    Topline corresponds to the original data encoded with 32-bit PCM. 
    }
    \vspace{-0.2cm}
    \label{tab:bitrate}
\end{table}

\vspace{-0.2cm}
\section{Examples}
\label{sec:examples}
\vspace{-0.2cm}

Alongside the core functionality of the library, we provide a set of illustrative examples. The goal of these examples is two-fold: (a) to illustrate the usage of particular components of the library, and (b) to serve as a starter code for a particular type of application. For instance, a probing example (Section~\ref{ss:probing}) can be adapted for better studying used representations, while discrete resynthesis (Section~\ref{ss:resynth}) could provide a starter code for an application operating on units (e.g., language modeling or a high-compression speech codec).

\begin{table*}[ht!]
    \centering
    \resizebox{0.8\textwidth}{!}{
    \small
    \begin{tabular}{l}
    \toprule
         \texttt{{\color{magenta} HE PASSES ABRUPTLY FROM PERSONS} OF ABRUPT ACID FROM WHICH HE PROCEEDS ARIGHT BY ...} \\
         \texttt{{\color{magenta} HE PASSES ABRUPTLY FROM PERSONS} AND CHARCOAL EACH ONE OF THE CHARCOAL ...}\\ 
         \texttt{{\color{magenta} HE PASSES ABRUPTLY FROM PERSONS} FEET AND TRAY TO A CONTENTION OF ASSOCIATION THAT ...} \\ 
         \bottomrule
    \end{tabular}}
    \caption{Three continuations of the same prompt (in pink), generated by the speech continuation example under different random seeds. Sampled from a language model trained trained on HuBERT-100 units.}
    \vspace{-0.2cm}
    \label{tab:continuation}
\end{table*}

\vspace{-0.1cm}
\subsection{Speaker Probing}\label{ss:probing}
A vibrant area of research studies properties of ``universal'' pre-trained representations, such as GLoVE~\cite{Pennington2014} and~BERT~\cite{Devlin2018}. Examples span from probing for linguistic properties~\cite{adi2016fine,ettinger2016probing,adi2017analysis,Conneau2018,hewitt2019structural} to discovering 
biases~\cite{Bolukbasi2016,Caliskan2017}.

In contrast, widely used pre-trained representations produced by HuBERT~\cite{Hsu2021} and wav2vec~2.0~\cite{Baevski2020} are relatively understudied. Few existing works include~\citep{Niekerk2021,Higy2021}.

We believe our library can provide a convenient tool for research in this area. Hence, as the first example, we include a probing experiment similar to the one proposed in~\citep{Niekerk2021,adi2019reverse}. We study whether the extracted representations contain speaker-specific information. In this example, we experiment with quantized and continuous representations provided by CPC and HuBERT. We randomly split LibriSpeech dev-clean utterances into train/test (90\%/10\%) sets\footnote{We have to create a new split as the standard one has disjoint sets of speakers, making this experiment impossible.} and train a two-layer Transformer for 5 epochs to predict a speaker's anonymized identifier, based on an utterance they produced. From the results reported in Table~\ref{tab:speaker_probing} we see that the continuous representations allow identifying speaker on hold-out utterances. In contrast, the quantization adds some speaker-invariance, justifying its use.

\vspace{-0.1cm}
\subsection{Speech Resynthesis}
\label{ss:resynth}
The next example is the discrete speech resynthesis, i.e., the speech audio $\rightarrow$ deduplicated units $\rightarrow$ speech audio pipeline. Fig.~\ref{fig:resynth_listing} illustrates how simple its implementation is with \texttt{textless-lib}.

The discrete resynthesis operation can be seen as a lossy compression of the speech. Indeed, if a sequence of $n$ units  (from a vocabulary $\mathcal{U}$) encodes a speech segment of length $l$, we straightforwardly obtain a lossy codec with bitrate $\frac{n}{l} \lceil \log_2 |\mathcal{U}| \rceil$ bits per second. Further, the token stream itself can be compressed using entropy encoding and, assuming a unigram 
token model, the compression rate becomes:
$-\frac{n}{l} \cdot \sum_{u \in \mathcal{U}} \mathbb{P}(u) \log_2 \mathbb{P}(u).$
In Table~\ref{tab:bitrate} we report compression rate/word error rate (WER) trade-off achievable with the HuBERT-derived unit systems, as a function of the vocabulary size. WER is calculated using the wav2vec 2.0-based \ac{ASR} w.r.t.\ and uses the ground-truth transcripts. To calculate the compression rate, the unigram token distribution was fitted on the transcript of LibriLight 6K dataset~\cite{Riviere2020}. 
From Table~\ref{tab:bitrate} we observe that discretized HuBERT representations have a strong potential for extreme speech compression~\citep{Polyak2021}.\footnote{In contrast to our setup, \citet{Polyak2021} worked with non-deduplicated streams, hence obtained different bitrates.} Our provided implementation reports the bitrate.

\vspace{-0.1cm}
\subsection{Speech Continuation}
Finally, we include a \texttt{textless-lib} re-implementation of the full GSLM speech continuation pipeline~\cite{Lakhotia2021}, as depicted in Figure~\ref{fig:pipeline}. Table~\ref{tab:continuation} presents \ac{ASR} transcripts of three different continuations of the same prompt, generated using different random seeds. We use a \textsc{large} wav2vec 2.0~\href{https://github.com/pytorch/fairseq/tree/master/examples/wav2vec\#wav2vec-20}{model}, trained on LibriSpeech-960h with CTC loss. Its decoder uses the standard KenLM 4-gram language model.
\vspace{-0.1cm}
\section{Discussion and Future Work}
\label{sec:dis}
\vspace{-0.1cm}

We introduced \texttt{textless-lib}, a Pytorch library aimed to advance research in textless modeling of spoken language, by simplifying textless processing and synthesizing spoken language. We described the main building blocks used to preprocess, quantize, and synthesize speech. To demonstrate the usability of the library, we provided three usage examples related to (i) representation probing, (ii) speech compression, and (iii) speech continuation. The proposed library greatly simplifies research in the textless spoken language processing, hence we believe it will be a handful not only for speech researchers but to the entire NLP community. 

As a future work for \texttt{textless-lib} we envision  improving performance of the existing building blocks, adding new example tasks (e.g., translation), extending the set of  provided pre-trained models, and introducing the possibility of training the different components.

\bibliography{anthology,custom}
\bibliographystyle{acl_natbib}

\end{document}